%% file: iclr2025_conference.tex
\documentclass{article} 
\usepackage{iclr2025_conference,times}

\input{math_commands.tex}

\usepackage{multirow}
\usepackage{listings}
\usepackage{hyperref}
\usepackage{url}
\usepackage{listings}
\usepackage{xcolor}
\usepackage{booktabs}
\usepackage{caption}
\usepackage{soul}
\usepackage{graphicx}
\usepackage{pifont}
\usepackage{enumitem}
\setlist[itemize]{leftmargin=*}
\setlist[enumerate]{leftmargin=*}
\newcommand{\cmark}{\ding{51}}%
\newcommand{\xmark}{\ding{55}}%

\title{Adapting Language-Specific LLMs to a Reasoning Model in One Day via Model Merging - An Open Recipe}

\author{Kunat Pipatanakul, Pittawat Taveekitworachai,\\
\textbf{Potsawee Manakul, and Kasima Tharnpipitchai}\\
SCB 10X R\&D\\
SCBX Group\\
Bangkok, Thailand \\
\texttt{\{kunat,pittawat,potsawee,kasima\}@scb10x.com}
}

\iclrfinalcopy
\begin{document}
\maketitle

\begin{abstract}
This paper investigates data selection and model merging methodologies aimed at incorporating advanced reasoning capabilities such as those of DeepSeek R1 into language-specific large language models (LLMs), with a particular focus on the Thai LLM.
Our goal is to enhance the reasoning capabilities of language-specific LLMs while maintaining their target language abilities.
DeepSeek R1 excels in reasoning but primarily benefits high-resource languages such as English and Chinese. However, low-resource languages remain underserved due to the dominance of English-centric training data and model optimizations, which limit performance in these languages. This limitation results in unreliable code-switching and diminished effectiveness on tasks in low-resource languages.
Meanwhile, local and regional LLM initiatives have attempted to bridge this gap by developing language-specific LLMs that focus on improving local linguistic fidelity.
We demonstrate that, with only publicly available datasets and a computational budget of \$120\footnote{4 H100 GPUs × 15 hours × \$2 per hour = \$120}, it is possible to enhance the reasoning capabilities of language-specific LLMs to match the level of DeepSeek R1, without compromising their performance on target language tasks.
This work releases the data, merge configurations, and model weights to promote the advancement of language-specific LLM initiatives.\footnote{Research artifacts: \url{https://huggingface.co/collections/scb10x/typhoon-r1-iclr-2025-sci-fm-artifacts-67e3adf1cc93cc8c42eaebb3}}

\end{abstract}

\section{Introduction}
\label{intro}

Recent advancements in large language models (LLMs) have demonstrated remarkable capabilities in complex reasoning tasks, particularly through innovations in scaling at test time and specialized training paradigms \citep{deepseekai2025deepseekr1incentivizingreasoningcapability}. Notable breakthroughs by models such as OpenAI o1 and DeepSeek R1 \citep{deepseekai2025deepseekr1incentivizingreasoningcapability} have established new standards in tackling reasoning challenges that were previously difficult for LLMs. However, these achievements primarily focus on high-resource languages, particularly English and Chinese, creating a significant gap in capabilities for low-resource languages.

The underlying foundation models, such as Llama \citep{grattafiori2024llama3herdmodels} and Qwen \citep{qwen2025qwen25technicalreport}, predominantly rely on training data in English and Chinese, leading to limitations when applied to low-resource languages. While these models may achieve impressive scores on certain low-resource language benchmarks, they frequently exhibit issues such as incorrect character usage and code-switching in practical applications, see example in Appendix~\ref{appendix:code_switching_example}. These problems become more pronounced during reasoning-focused fine-tuning and reinforcement learning \citep{qwq-32b-preview}, where both the query language and solution paths are optimized in high-resource languages.

Several local and regional LLM initiatives, including Typhoon \citep{pipatanakul2024typhoon2familyopen}, Sailor \citep{dou2024sailoropenlanguagemodels}, EuroLLM \citep{martins2024eurollmmultilinguallanguagemodels}, Aya \citep{ayamodelinstructionfinetuned}, Sea-lion \citep{sea_lion_2024}, and SeaLLM \citep{nguyen2024seallmslargelanguage}, have attempted to address these limitations through continuous pretraining and post-training approaches tailored to specific target languages. 
However, a data-centric approach to adapting reasoning capabilities for low-resource languages through knowledge distillation presents two main challenges: (1) most reasoning models do not disclose their data recipes, creating uncertainties about the impact of source prompts and reasoning trace quality on model performance, an underexplored area; (2) scaling data-centric approaches for teaching reasoning capabilities requires substantial computational resources. For example, adapting DeepSeek R1 70B involved 600K examples for distillation and 200K for general SFT, totaling 800K examples, far exceeding the 17K used in Sky-T1 \citep{sky_t1_2025}, Bespoke-Stratos \citep{bespoke_stratos}, and similar academic efforts, which cost several thousand dollars per run.

In parallel to the aforementioned efforts, model merging has shown promise in combining the weights of multiple models with different specializations, resulting in new models that improve performance across multiple tasks without requiring additional training \citep{siriwardhana2024domainadaptationllama370binstructcontinual, lu2024finetuninglargelanguagemodels, akiba2025evolutionaryoptimizationmodelmerging}. In practice, we can frame the problem of combining two models—one with reasoning capability and one specialized in language—as a merging task between models with different specializations. This involves treating the reasoning model as one specialized component and the language-focused model as another, then integrating them via merging.

In this work, we examine techniques for data selection, augmentation using only publicly available datasets, and optimize model merging ratio to integrate DeepSeek R1’s reasoning capabilities into a 70B Thai language model. Our goal is to enhance the reasoning capabilities of language-specific models while maintaining strong performance in the target language. We demonstrate that by leveraging publicly available datasets and an affordable computational budget, it is possible to significantly improve language-specific models to match DeepSeek R1's reasoning capabilities without compromising general language tasks performance.

\section{Methodology}
\label{method}

\begin{figure}[!ht]
\begin{center}
\includegraphics[width=110mm]{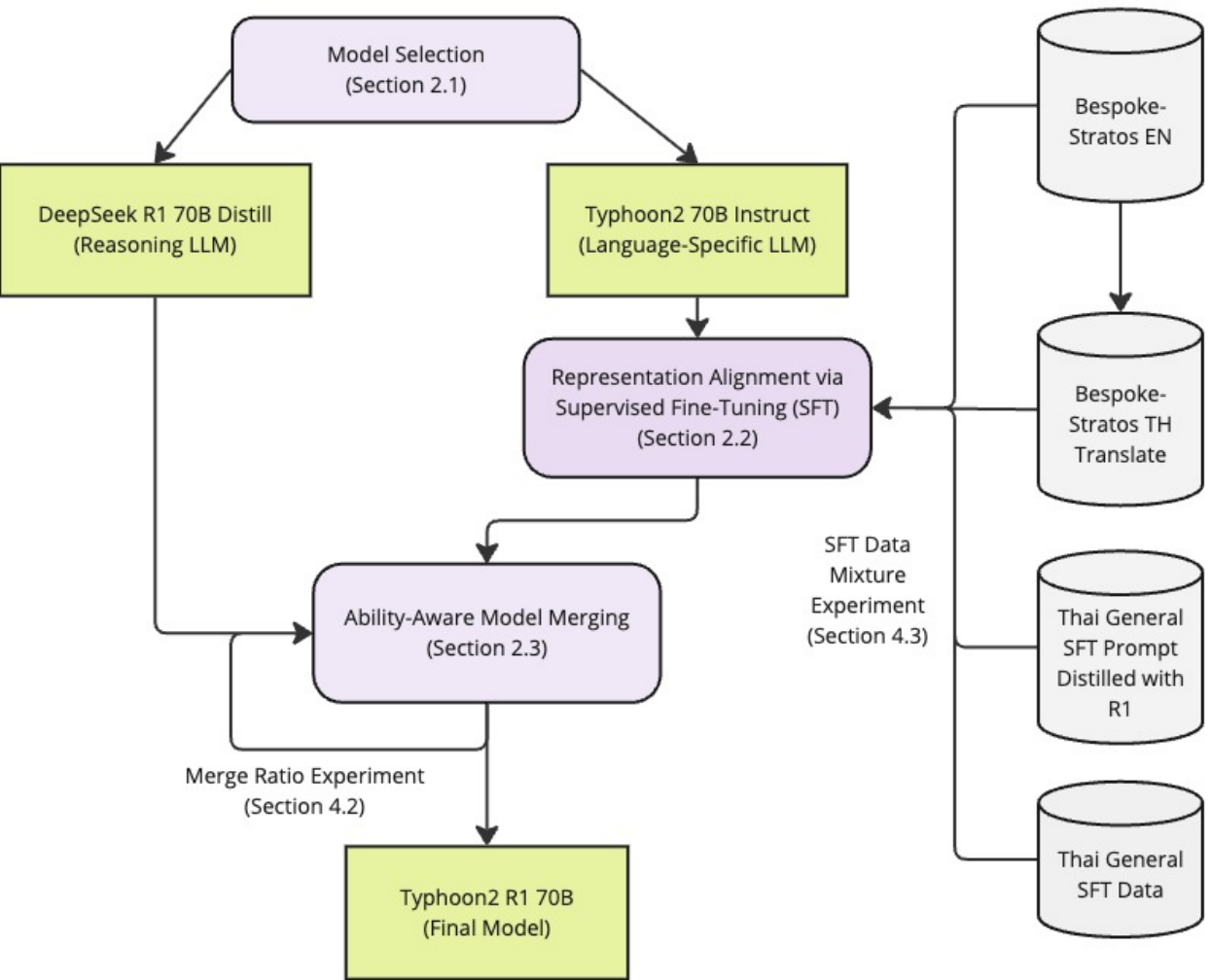}
\end{center}
\caption{Overview of our Typhoon2 R1 70B recipe}
\label{fig:method}
\end{figure}

In this section, we first explain the motivation and process behind our approach. First, we select two specialized LLMs--one proficient in a target low--resource language (e.g., Thai) and the other specialized in long-thought reasoning (Section~\ref{method_modelselection}). After selecting the models, we employ a two-stage procedure:

\begin{enumerate}
    \item Representation Alignment via Supervised Fine-Tuning (SFT), described in Section~\ref{method_sft})
    \item Ability-Aware Model Merging, described in Section~\ref{method_modelmerging})
\end{enumerate}

\subsection{Model Selection}
\label{method_modelselection}

To integrate advanced reasoning with strong language capabilities in a low-resource setting, we begin by selecting two LLMs that share the same foundational architecture. This architectural compatibility facilitates effective alignment and merging of learned parameters. Specifically, we choose one language-specific model specialized in Thai and another long-thought reasoning model. 

For our approach, we choose Typhoon2 70B Instruct \citep{pipatanakul2024typhoon2familyopen}, a Thai-specialized model derived from Llama 3.1 70B. This model has undergone continuous pretraining (CPT) and alignment using a Thai-focused dataset to enhance its performance in Thai. Additionally, we incorporate DeepSeek R1 70B Distill \citep{deepseekai2025deepseekr1incentivizingreasoningcapability}, a reasoning-focused model fine-tuned from Llama 3.3 70B and SFT from an 600K-instance distilled reasoning trace dataset generated by DeepSeek R1 + 200K general SFT dataset.

Both models belong to the Llama 3.1 70B family \citep{grattafiori2024llama3herdmodels}, sharing a common pre-trained backbone. This shared foundation ensures well-aligned internal representations, improving the effectiveness of subsequent merging steps.

\subsection{Representation Alignment via Supervised Fine-Tuning (SFT)}
\label{method_sft}

To facilitate effective model merging, we first align the internal representations of the language-specific and reasoning-focused LLMs through supervised fine-tuning. The goal of this step is to ensure that the two models develop a similar representation space, enabling smoother integration in the subsequent merging phase.

For this purpose, we construct a training dataset that promotes alignment between language and reasoning capabilities. Our foundation is Bespoke-Stratos \citep{bespoke_stratos}, a carefully curated reasoning distillation dataset comprising 17,000 examples generated by DeepSeek R1. These examples have demonstrated state-of-the-art (SOTA) performance in reasoning tasks when fine-tuned (SFT) directly on Qwen2.5-32B-Instruct \citep{qwen2025qwen25technicalreport}\footnote{https://www.bespokelabs.ai/blog/bespoke-stratos-the-unreasonable-effectiveness-of-reasoning-distillation}. However, since Bespoke-Stratos is primarily in English, we adapt it for our low-resource setting by translating the question and solution components into Thai while retaining the original English reasoning traces.
Intuitively, this bilingual setup aims to teach model to align Thai question-solution pairs to the original English long-form reasoning traces. This approach is inspired by a work on multilingual CoT \citep{shi2022languagemodelsmultilingualchainofthought}.

In addition to this translated dataset, we explore additional datasets to assess the impact of each data type on the final merged model’s performance, including (i) Distillation of DeepSeek R1 long-thought reasoning examples on general Thai prompts; (ii) General Thai instruction-tuning dataset.


\subsection{Ability-Aware Model Merging}
\label{method_modelmerging}
After fine-tuning the language-specific model, we merge its parameters with those of the long-thought reasoning model using a merge operation, specifically, we adopt the method proposed by \cite{yu2024languagemodelssupermario}. Additionally, we optimize the merge ratio based on the hypothesis that different layers contribute uniquely to distinct capabilities, as observed in early empirical experiments. 
Intuitively:

\label{method_merge_intuitive}
\begin{itemize}
    \item \textbf{Early to middle layers} are assigned a higher weight from the long-thought reasoning model, as these layers primarily handle comprehension and abstract thinking.

    \item \textbf{Later layers} are more influenced by the language-specific model, as they play a crucial role in fluent generation in the output.
\end{itemize}

The objective is to create a merged model that excels in both reasoning and Thai capabilities.

\section{Experimental Setup}
\subsection{Base Models and Training Configuration}
We select Typhoon2 70B Instruct and DeepSeek R1 70B Distill as the base models for all experiments. SFT is applied to Typhoon2 70B, and we merge DeepSeek R1 70B with Typhoon2+SFT.

All models are trained using LoRA with a rank of 32 and $\alpha$ of 16, employing a cosine learning rate schedule with a peak learning rate of 4e-4 over a single training epoch. To enhance computational efficiency, we utilize sequence packing with a maximum length of 16,384, along with Liger kernels \citep{hsu2024ligerkernelefficienttriton}, FlashAttention-2 \citep{dao2023flashattention2fasterattentionbetter}, and DeepSpeed ZeRO-3. \citep{rajbhandari2020zeromemoryoptimizationstraining} Each experiment is conducted on 4×H100 GPUs for a maximum of 15 hours. Training is performed using axolotl\footnote{\url{https://github.com/axolotl-ai-cloud/axolotl/}}. Model merging is performed using Mergekit \citep{goddard2025arceesmergekittoolkitmerging}.

\subsection{Evaluation}
\label{evaluation}
Evaluation is conducted on two aspects: (i) reasoning capability and (ii) performance on language tasks. \textit{First}, the reasoning capability is assessed using standard benchmarks commonly employed for evaluating reasoning models, such as AIME 2024 and MATH-500 for the mathematics domain, and LiveCodeBench for the coding domain based on evaluation in Sky T1 \citep{sky_t1_2025}. We translate the math and code evaluation query to Thai using GPT-4o to assess the performance in Thai on reasoning tasks.
\textit{Second}, for languages tasks, we focus on assessing general instruction-following performance and the usability of LLMs in Thai. Which consist of IFEval, MT-Bench and Language accuracy.
We also add Think accuracy, which represents how well the model can correctly output its thoughts before responding.
All evaluation datasets are as follows:
\begin{itemize}
    \item \textbf{IFEval:} We use IFEval \citep{ifeval} to evaluate instruction-following capabilities based on verifiable instructions and predefined test cases, measuring accuracy as the adherence metric. Alongside the standard (English) IFEval, we employ the Thai translated version \texttt{scb10x/ifeval-th}\footnote{\url{https://huggingface.co/datasets/scb10x/ifeval-th}}. 

    \item \textbf{MT-Bench:} We adopt MT-Bench, an LLM-as-a-judge framework, to assess responses based on correctness, fluency, and adherence to instructions. For Thai, we use Thai MT-Bench, proposed by ThaiLLMLeaderboard \citep{thaillm-leaderboard,mtbenchth}, while the English version follows LMSYS~\citep{mtbench}.

    \item \textbf{Language Accuracy:} We use code-switching metrics based on \cite{pipatanakul2024typhoon2familyopen}, inspired by IFEval \citep{ifeval}. Lang-acc measures the model ability to respond in the query language while adhering to natural linguistic conventions. Specifically, it must comply with two criteria:
    (1) Valid responses should contain only Thai and English, excluding Chinese or Russian, as these represent the languages commonly used by Thai speakers in daily life.
    (2) English usage must follow native conventions, with the total number of English characters being fewer than Thai characters.
    Evaluations are conducted using 500 sampled instructions from \texttt{airesearch/WangchanThaiInstruct}~\citep{wangchaninstruct}, tested at $T=0.7$ to ensure consistency in a non-greedy setting. More details are provided in~\ref{appendix:language_acc}.

    \item \textbf{Think Accuracy:} Inspired by R1 think patterns, think accuracy quantifies instances where the LLM generates structured ``thinking trace`` responses across the evaluation dataset. We measure think accuracy by assessing all responses obtained from other evaluations in this setup and aggregating their tendency to think before answering by checking the presence of the `\texttt{$</$think$>$}` token and analyzing the content between think tokens. More details are provided in Appendix~\ref{appendix:think_acc}.

    \item \textbf{MATH500:} We use MATH-500 \citep{lightman2023letsverifystepstep} subset of the MATH benchmark \cite{hendrycks2021measuringmathematicalproblemsolving}, a dataset consisting of 500 problems to evaluate math reasoning ability. Additionally, we include a translated set, bringing the total to 1,000 problems.

    \item \textbf{AIME 2024:} Also to evaluate math reasoning ability, we use AIME 2024\footnote{\url{https://huggingface.co/datasets/HuggingFaceH4/aime_2024}}, which is derived from 2024 USA Math Olympiad (AIME). The evaluation consists of 30 challenging problems. Additionally, we translate the dataset into Thai, resulting in a total of 60 problems.

    \item \textbf{LiveCodeBench:} We evaluate code reasoning ability using LiveCodeBench \citep{jain2024livecodebench}, which collects coding problems from three competitive programming platforms: LeetCode, AtCoder, and CodeForces. We use the release\_v2 version, which contains 511 problems, along with an equally translated Thai version.
    
\end{itemize}

We also report `Average' columns, which represent the average results from all evaluations. To ensure equal weighting, we multiply the MT-Bench score by 10 and use a simple arithmetic mean.

\section{Results and Discussion}



\subsection{The English-Thai Performance Gap}
\label{section:gap}
First, we examine the performance of our two base models on language and reasoning tasks to understand the performance gap between them.

As shown in \ref{tab:baseline}, while DeepSeek R1 70B Distill performs well in reasoning tasks such as AIME and MATH500, its performance in IFEval is slightly worse but still acceptable. However, in MT-Bench-TH and language accuracy tasks, it falls significantly behind language-specific models such as Typhoon2 70B Instruct.
On the other hand, Typhoon2 70B Instruct struggles with most reasoning tasks, achieving only 10\% accuracy in AIME and trailing DeepSeek R1 70B Distill by more than 20\% in MATH500.

\begin{table}[!ht]
\tabcolsep=1mm
    \caption{Performance comparison between Typhoon2 70B Instruct and DeepSeek R1 70B Distill, Showing that Typhoon2 have stronger language task performance, while DeepSeek has stronger reasoning performance. However, neither model compensates for its weakness.}
    \centering
    \small
    \begin{tabular}{l|ccccccccccccc}
    \toprule
         \multirow{2}{*}{Experiment} &\multicolumn{2}{c}{IFEval} &\multicolumn{2}{c}{MT-Bench}  &\multicolumn{2}{c}{Response Acc}  &\multicolumn{2}{c}{AIME} &\multicolumn{2}{c}{MATH500}  &\multicolumn{2}{c}{LCB}&\multirow{2}{*}{Avg.}  \\ 
        &EN &TH &EN &TH &Lang &Think &EN &TH &EN &TH &EN &TH\\
        \midrule
        \texttt{Typhoon2 70B} &\textbf{88.7} &\textbf{81.4} &8.856 &\textbf{7.362} &\textbf{98.8} &0.0 &10.0 &3.3 &66.2 &60.9 &39.9 &36.4& 54.0\\
        \texttt{Deepseek R1 70B}  &85.7 &74.3 &\textbf{8.939} &6.329 &19.0 &\textbf{84.2} &\textbf{63.3} &\textbf{40.0} &\textbf{88.4} &\textbf{78.7} &\textbf{64.7} &\textbf{62.8}& \textbf{67.8}\\
    \bottomrule
    \end{tabular}
    \label{tab:baseline}
\end{table}



\subsection{Merge Ratio}
\label{section:merge}
Based on the performance gap shown in Section~\ref{section:gap}, our goal is to combine the strengths of both models. To achieve this, we investigate model merging. 
In this section, We design two experiments to explore the merge ratio based on the intuition described in Section~\ref{method_merge_intuitive}. We employ a merging strategy based on the dare-linear method \citep{yu2024languagemodelssupermario} and constrain the search space by optimizing only the mixing ratios of two models: Typhoon+SFT-v1 and DeepSeek R1 70B Distill.

To conduct this experiment, we fine-tuned the SFT-v1 model, which includes Bespoke-Stratos (English) and a 2K Thai translation of Bespoke-Stratos. The details of the dataset used for SFT-v1 are provided in Table~\ref{tab:sft_data}.

In this case, we try to confirm our hypothesis in Section~\ref{method_merge_intuitive} with this two questions

\subsubsection{Which model should be assigned a higher ratio in the final merge to best preserve its reasoning capabilities?}
\label{exp_merge_1}
In this experiment, we maintained a fixed ratio of merged to single constraints across all model layers. Specifically, we applied a 25\%|75\% merge ratio to each model combination. This experiment is represented by configurations M1 (More Typhoon) and M2 (More DeepSeek).

\begin{table}[!ht]
    \caption{Merge config for question~\ref{exp_merge_1}, where DS-R @ K represents the DeepSeek ratio (DS-R) at layer K}
    \centering
    \small
    \tabcolsep=0.8mm
    \begin{tabular}{r|c}
    \toprule
        Merge Config & DS-R @ Layer 0 - Layer 80 \\ 
    \midrule
        M1 (More Typhoon) & 25\% \\
        M2 (More DeepSeek) & 75\% \\
    \bottomrule
    \end{tabular}
    \label{tab:merge_ratio_1}
\end{table}

As shown in Table \ref{tab:merge_result_1}, our findings indicate that a high ratio of DeepSeek R1 70B Distill in M2 improves performance across all evaluation metrics including reasoning tasks. Even within 3\%\footnote{An average score of 64.90\% vs. 66.32\% on AIME, MATH500 and LiveCodeBench.} of original DeepSeek R1 70B Distill.
However, there is still a degradation in the \textit{Language Accuracy} task. This outcome aligns with expectations, as DeepSeek R1 70B Distill struggles to generate Thai reliably.

\begin{table}[!ht]
\tabcolsep=1.3mm
    \caption{Comparison between the merged models: M1 (More Typhoon) and M2 (More DeepSeek), showing that M2 performs better overall but still exhibits degradation in language accuracy.}
    \centering
    \small
    \begin{tabular}{l|ccccccccccccc}
    \toprule
         \multirow{2}{*}{Experiment} &\multicolumn{2}{c}{IFEval} &\multicolumn{2}{c}{MT-Bench}  &\multicolumn{2}{c}{Response Acc} &\multicolumn{2}{c}{AIME} &\multicolumn{2}{c}{MATH500}  &\multicolumn{2}{c}{LCB} & \multirow{2}{*}{Avg.} \\ 
        &EN &TH &EN &TH &Lang &Think  &EN &TH &EN &TH &EN &TH\\
        \midrule
        M1 &57.4 &58.2 &7.728 &6.412 &\textbf{86.4} & 96.6 & 26.6 & 26.6 & 82.4 & 78.5 & 43.8 & 44.6 & 61.9\\
        M2 &\textbf{86.9} &\textbf{76.0} &\textbf{8.606} &\textbf{6.950} &59.8 & \textbf{100.0} &\textbf{46.6} &\textbf{50.0} &\textbf{89.8} &\textbf{83.7} & \textbf{58.3} &\textbf{61.0} &\textbf{72.3}\\
        \midrule
        Deepseek R1 70B  &85.7 &74.3 &8.939 &6.329 &19.0 &84.2 &63.3 &40.0 &88.4 &78.7 &64.7 &62.8& 67.8\\
    \bottomrule
    \end{tabular}
    \label{tab:merge_result_1}
\end{table}

\subsubsection{How does increasing Typhoon’s contribution in the later layers enhance language-specific performance?}
\label{exp_merge_2}
After finding that M2 performs better on reasoning tasks and many language tasks but has lower language response accuracy, which may reduce its usefulness for end users, we explore ways to improve language response accuracy while preserving reasoning capability. To achieve this, we allocate a higher ratio to Typhoon in the later layers. This experiment is represented by M2 and M3.

\begin{table}[!ht]
    \tabcolsep=1.2mm
    \caption{Merge config for question~\ref{exp_merge_2}, where DS-R @ K represents the DeepSeek ratio (DS-R) at layer K}
    \centering
    \small
    \begin{tabular}{l|c|c}
    \toprule
        Merge Config & DS-R @ Layer 0-53 & DS-R @ Layer 53-80 \\ 
        \midrule
        M2 (Constraint ratio) & 75\% & 75\%\\
        M3 (More Typhoon in later layer) & 75\% & 75\% linearly decrease to 12.5\% \\
    \bottomrule
    \end{tabular}
    \label{tab:merge_ratio_2}
\end{table}

As shown in Table~\ref{tab:merge_result_2}, reducing the contribution of DS-R @ layer 80 to 12.5\% and increasing Typhoon’s contribution to 87.5\% increase language accuracy to 87.6\%, representing an improvement of over 25\%. Additionally, this adjustment increases MT-Bench TH scores, indicating improved Thai language performance. Overall, the language benchmark performance gap decreases from 12.2\%\footnote{An average score of 75.65\% vs. 86.21\% on IFEval, MT-Bench, and Language accuracy.} to 6.8\%\footnote{An average score of 80.35\% vs. 86.21\% on IFEval, MT-Bench, and Language accuracy.}. Meanwhile, performance in reasoning tasks remains comparable to the M2(Constraint ratio) configuration.

Based on these findings, we ultimately select M3 as our final merge configuration.

\begin{table}[!ht]
\tabcolsep=1.4mm
    \caption{Performance comparison between the merged model with M2(Constraint ratio) and M3(More Typhoon in the later layer), showing that M3 improves language accuracy and enhances overall performance.}
    \centering
    \small
    \begin{tabular}{l|ccccccccccccc}
    \toprule
         \multirow{2}{*}{Experiment} &\multicolumn{2}{c}{IFEval} &\multicolumn{2}{c}{MT-Bench}  &\multicolumn{2}{c}{Response Acc} &\multicolumn{2}{c}{AIME} &\multicolumn{2}{c}{MATH500}  &\multicolumn{2}{c}{LCB} & \multirow{2}{*}{Avg.} \\ 
        &EN &TH &EN &TH &Lang &Think  &EN &TH &EN &TH &EN &TH\\
        \midrule
        M2 &\textbf{86.9} &\textbf{76.0} &\textbf{8.606} &6.950 &59.8 & \textbf{100.0} &\textbf{46.6} & \textbf{50.0} &89.8 &\textbf{83.7} &\textbf{58.3} &\textbf{61.0} & 72.3 \\
        M3 &82.9 &75.7 &8.390 &\textbf{7.164} &\textbf{87.6} & \textbf{100.0} &\textbf{46.6} &40.0 &\textbf{90.0} &81.9 &55.9 &58.5& \textbf{72.9} \\
        \midrule
        Typhoon2 70B &88.7 &81.4 &8.856 &7.362 &98.8 &0.0 &10.0 &3.3 &66.2 &60.9 &39.9 &36.4& 54.0\\
    \bottomrule
    \end{tabular}
    \label{tab:merge_result_2}
\end{table}

The merge configuration for Mergekit \cite{goddard2025arceesmergekittoolkitmerging} is provided in Appendix \ref{appendix:merge_config}.



\subsection{Supervised Fine Tuning (SFT): Data Mixture}
\label{section:sft}
After identifying a merge configuration that effectively combines the abilities of two models in Section~\ref{section:merge}, we focus on optimizing the data mixture for the SFT model to enhance alignment before merging, ultimately improving end-to-end performance.

In this section, we explore the impact of the SFT dataset on overall model performance by addressing the following key dataset considerations

\begin{enumerate}
    \item \textbf{Does increasing the data mixture of Thai to 30\% improve performance compared to 10\%?} - We investigate the impact of Thai-English data proportions, we add an additional 4.5k Thai translation examples based on translation of Bespoke-Stratos as in Section~\ref{method_sft}, which increase the Thai language ratio from 10\% to 30\%.

    \item \textbf{Does adding distilled reasoning traces on general Thai queries improve performance?} - We hypothesize that Bespoke-Stratos primarily covers math, code, and puzzle domains, lacking diversity in instruction-following tasks. Does adding general-domain distillation with long-form reasoning improve performance? To test this hypothesis, we sample 1,000 prompts from the Thai general instruction dataset \texttt{Suraponn/thai\_instruction\_sft}\footnote{\url{https://huggingface.co/datasets/Suraponn/thai_instruction_sft}}, distill responses using DeepSeek R1, and apply rejection sampling to exclude non-Thai solutions, retaining approximately 50\% of the samples. The final dataset consists of 500 examples.

    \item \textbf{Does adding a general instruction dataset improve performance?} - We hypothesize that adding a general instruction dataset might improve dataset diversity and help prevent catastrophic forgetting. To investigate this, we incorporate 10,000 general instruction examples. For English, we use Capybara\footnote{\url{https://huggingface.co/datasets/LDJnr/Capybara}}, and for Thai, we use \texttt{Suraponn/thai\_instruction\_sft}, following its usage in Typhoon 2 \citep{pipatanakul2024typhoon2familyopen}. Each dataset is subsampled to 10,000 examples to maintain balance.
\end{enumerate}

We construct dataset based on the above question which can is summarized in Table \ref{tab:sft_data}.
The SFT model is later merged using the M3 configuration before evaluation, as we try to optimize for end-to-end performance.

\begin{table}[!ht]
    \tabcolsep=0.6mm
    \caption{A summary of the SFT data configurations used in our SFT: data mixture experiment.}
    \centering
    \small
    \begin{tabular}{lllcccc}
    \toprule
        Dataset & Language &\#Examples & SFT-V1 & SFT-V2 & SFT-V3 & SFT-V4  \\ 
        \midrule
        Bespoke-Stratos (Original) & EN & 17K & \cmark &\cmark & \cmark & \cmark \\
        Bespoke-Stratos TH Translate (Small) & TH & 2K & \cmark &\xmark & \xmark & \xmark \\
        Bespoke-Stratos TH Translate (Large) & TH & 6.5K & \xmark &\cmark & \cmark & \cmark \\
        Deepseek R1 Distill \texttt{thai\_instruction\_sft} & TH & 0.5K & \xmark &\xmark & \cmark & \cmark \\
        Capybara (Original) & EN & 10K & \xmark &\xmark & \xmark & \cmark \\        \texttt{thai\_instruction\_sft} (Original) & TH & 10K & \xmark &\xmark & \xmark & \cmark \\
    \bottomrule
    \end{tabular}
    \label{tab:sft_data}
\end{table}

\textbf{Results:}
\label{exp_sft_result}
As shown in Table \ref{tab:sft_results}, we began with SFT-v1 + M3 as our baseline. First, we add 4.5k Thai translations from the Bespoke-Stratos dataset, in SFT-v2. This resulted in a slight improvement, primarily in general performance for both English and Thai. Next, we incorporated 500 distilled Thai responses from DeepSeek R1, which mainly enhanced language-accuracy performance. We hypothesize that the model benefited from greater generalization on general task due to increased diversity in prompts. Further, we experimented with adding general instruction-domain data. The results were mixed, with performance remaining comparable to SFT-v3. We suspect this may be due to the instruction dataset quality. Additionally, the general instruction data spans multiple dimensions, suggesting that further investigation is needed to understand its effects comprehensively.

Based on these findings, we use SFT-v3 as our final dataset mixture.

\begin{table}[!ht]
\tabcolsep=0.8mm
    \caption{Performance comparison of each SFT mixture. Result in Section~\ref{section:sft}}
    \centering
    \fontsize{8}{8}\selectfont
    \begin{tabular}{l|ccccccccccccc}
    \toprule
         \multirow{2}{*}{Experiment} &\multicolumn{2}{c}{IFEval} &\multicolumn{2}{c}{MT-Bench}  &\multicolumn{2}{c}{Response Acc} &\multicolumn{2}{c}{AIME} &\multicolumn{2}{c}{MATH500}  &\multicolumn{2}{c}{LCB} &\multirow{2}{*}{Avg.}\\ 
         &EN &TH &EN &TH &Lang &Think &EN &TH &EN &TH &TH & EN\\
        \midrule 
        SFT-v1 + M3 &82.9 &75.7  &8.390 &7.164 &87.6 &\textbf{100.0} &46.6 &40.0 &90.0 &81.9 &55.9 &58.5 &72.9\\
        +Add 4.5k TH translation (SFT-v2) &83.5 &\textbf{78.6}  &8.725 &7.082 &89.4 &99.9 &60.0 &\textbf{50.0} &\textbf{91.6} &82.1 &59.6 &\textbf{61.4} & 76.1\\
        +Distil 500 TH general thought (SFT-v3) &\textbf{85.1} &75.9 &\textbf{8.843} &\textbf{7.181} &\textbf{96.0} &99.9 &\textbf{63.3} &46.6 &90.4 &83.5 &\textbf{60.0} &57.3 &\textbf{76.5}\\
        +General Instruction (SFT-v4) &77.8 &77.8  &8.806 &6.939 &93.2 &99.7 &43.3 &46.6 &89.8 &\textbf{85.7} &53.8 &56.1 &73.4\\
    \bottomrule
    \end{tabular}
    \label{tab:sft_results}
\end{table}



\subsection{Does Directly Merging the Original Model Work?}
\label{exp_merge_only_work?}
Based on both merge configuration and SFT dataset mixture in Section~\ref{section:merge} and \ref{section:sft} we also validate that whether merging alone, without any SFT, is sufficient for the model to function properly. In this experiment, we compare our best model (Typhoon2+SFT-v3+M3) with a directly merged version (Typhoon2+M3), skipping SFT entirely.
Our results in Table \ref{tab:directly_merged} suggest that direct merging may not be effective, as it results in lower performance across all benchmarks.

\begin{table}[!ht]
\tabcolsep=1.3mm
    \caption{Performance comparison between our best model(Typhoon2+SFT-v3+M3) and direct merging(Typhoon2+M3).}
    \centering
    \small
    \begin{tabular}{l|ccccccccccccc}
    \toprule
         \multirow{2}{*}{Experiment} &\multicolumn{2}{c}{IFEval} &\multicolumn{2}{c}{MT-Bench}  &\multicolumn{2}{c}{Response Acc}  &\multicolumn{2}{c}{AIME} &\multicolumn{2}{c}{MATH500}  &\multicolumn{2}{c}{LCB} & \multirow{2}{*}{Avg.}  \\ 
        &EN &TH &EN &TH &Lang &Think &EN &TH &EN &TH &EN &TH\\
        \midrule 
        Typhoon2+M3 &77.0 &58.6 &8.581 &5.835 &90.8 &65.0 &46.6 &20.0 &88.2 &67.9 &\textbf{61.0} &47.3 &63.9\\
        Best Model &\textbf{85.1} &\textbf{75.9} &\textbf{8.843} &\textbf{7.181} &\textbf{96.0} &\textbf{99.9} &\textbf{63.3} &\textbf{46.6} &\textbf{90.4} &\textbf{83.5} &60.0 &\textbf{57.3} &\textbf{76.5}\\
    \bottomrule
    \end{tabular}
    \label{tab:directly_merged}
\end{table}

\subsection{Does SFT only Model Work?}
After verifying that the merged-only model does not work in Section~\ref{exp_merge_only_work?}, we also evaluate whether SFT alone, is effective. We set up the experiment in the same way as in Section~\ref{exp_merge_only_work?}. Specifically, we compare the best model (Typhoon2+SFT-v3+M3) with a directly fine-tuned version (Typhoon2+SFT-v3) without merging.

Although there are examples of SFT improving performance in high-resource languages \citep{sky_t1_2025, bespoke_stratos}, our results in Table~\ref{tab:directly_merged} suggest that direct SFT alone is not effective. It results in lower performance across all benchmarks, which may be due to the limitations of language-specific LLM capabilities, the use of LoRA in our setup, or other factors—an aspect left for future work.

\begin{table}[!ht]
\tabcolsep=1.3mm
    \caption{Performance comparison between our best model(Typhoon2+SFT-v3+M3) and direct SFT(Typhoon2+SFT-v3).}
    \centering
    \small
    \begin{tabular}{l|ccccccccccccc}
    \toprule
         \multirow{2}{*}{Experiment} &\multicolumn{2}{c}{IFEval} &\multicolumn{2}{c}{MT-Bench}  &\multicolumn{2}{c}{Response Acc}  &\multicolumn{2}{c}{AIME} &\multicolumn{2}{c}{MATH500}  &\multicolumn{2}{c}{LCB} & \multirow{2}{*}{Avg.}  \\ 
        &EN &TH &EN &TH &Lang &Think &EN &TH &EN &TH &EN &TH\\
        \midrule 
        Typhoon2+SFT-v3 &70.3 &60.9 &7.868 &6.412 &\textbf{98.6} &97.7	&10.0 &16.6 &72.8 &67.9	&35.8 &34.6 &59.0\\
        Best Model &\textbf{85.1} &\textbf{75.9}  &\textbf{8.843} &\textbf{7.181} &96.0 &\textbf{99.9} &\textbf{63.3} &\textbf{46.6} &\textbf{90.4} &\textbf{83.5} &60.0 &\textbf{57.3} &\textbf{76.5}\\
    \bottomrule
    \end{tabular}
    \label{tab:sft_only}
\end{table}

\subsection{Final Model}
\label{section:final_model}

Based on the combination of all experiments in this work, we found that our best model, which we call \textbf{Typhoon2-R1-70B}, demonstrates the feasibility of leveraging model merging to combine the reasoning ability of DeepSeek R1 70B Distill with the Thai language proficiency of Typhoon2 70B Instruct. The results, presented in Table~\ref{tab:final_result}, suggest that Typhoon2-R1-70B achieves performance within approximately 4\%\footnote{An average score of 83.44\% vs. 86.21\% on IFEval, MT-Bench, and Language Accuracy.} of Typhoon2 70B Instruct on language tasks and comparable\footnote{An average score of 66.85\% vs 66.31\% on AIME, MATH500, and LiveCodeBench.} on reasoning tasks. Additionally, it boosts average across all tasks performance by 41.6\% over Typhoon2 70B Instruct and by 12.8\% over DeepSeek R1 70B Distill.


\begin{table}[!ht]
\tabcolsep=0.5mm
    \caption{Performance comparison of Typhoon2 70B Instruct, Typhoon2 R1 70B (Best Model), and DeepSeek R1 70B Distill shows that we can combine the performance of two models into one using SFT and model merging.}
    \centering
    \small
    \begin{tabular}{l|ccccccccccccc}
    \toprule
         \multirow{2}{*}{Experiment} &\multicolumn{2}{c}{IFEval} &\multicolumn{2}{c}{MT-Bench}  &\multicolumn{2}{c}{Response Acc}  &\multicolumn{2}{c}{AIME} &\multicolumn{2}{c}{MATH500}  &\multicolumn{2}{c}{LCB}& \multirow{2}{*}{Avg.}  \\ 
        &EN &TH &EN &TH &Lang &Think &EN &TH &EN &TH &EN &TH\\
        \midrule
        Typhoon2 70B Instruct &\textbf{88.7} &\textbf{81.4} &\textbf{8.856} &\textbf{7.362} &\textbf{98.8} &0.0 &10.0 &3.3 &66.2 &60.9 &39.9 &36.4 &54.0 \\
        Typhoon2-R1-70B(Best Model) &85.1 &75.9  &8.843 &7.181 &96.0 &\textbf{99.9} &\textbf{63.3} &\textbf{46.6} &\textbf{90.4} &\textbf{83.5} &60.0 &57.3 &\textbf{76.5}\\
        Deepseek R1 70B &85.7 &74.3 &8.939 &6.329 &19.0 &84.2 &\textbf{63.3} &40.0 &88.4 &78.7& \textbf{64.7} &\textbf{62.8} &67.8 \\
    \bottomrule
    \end{tabular}
    \label{tab:final_result}
\end{table}

We also show additional sample responses from the model in Appendix~\ref{appendix:example_from_model}.

\subsection{Additional Model \& Additional Language}

Based on our final model configuration (Section~\ref{section:final_model}), we investigate whether our method can be transferred to another model. To validate this, we design an experiment applying our approach to South-east Asia (SEA) language-specific model that supports Thai: Sealion v3 70B Instruct \citep{sea_lion_2024}\footnote{\url{https://huggingface.co/aisingapore/llama3.1-70b-cpt-sea-lionv3-instruct}}. 

We apply our final recipe to Sealion 70B to ensure that the method is transferable between language-specific LLMs. As shown in \ref{tab:sealion}, we find that our method successfully transfers the reasoning capability of DeepSeek R1 70B to the Sealion model despite differences in the CPT and SFT recipes, similar to its effect on Typhoon. Additionally, it preserves comparable language performance.

In theory, our approach relies solely on translating the English reasoning dataset and target-language prompts. As a result, it should be adaptable to any language for which a language-specific model of the same reasoning model size and pretraining architecture is available. However, verifying this across additional languages is left for future work.

\begin{table}[!ht]
\tabcolsep=0.9mm
    \caption{Performance comparison of Sealion 70B Instruct, Sealion 70B Instruct+SFT-v3+M3 (Best recipe), and DeepSeek R1 70B Distill demonstrates that this recipe can be transferred between different CPT/SFT recipes of language-specific LLMs.}
    \centering
    \small
    \begin{tabular}{l|ccccccccccccc}
    \toprule
         \multirow{2}{*}{Experiment} &\multicolumn{2}{c}{IFEval} &\multicolumn{2}{c}{MT-Bench}  &\multicolumn{2}{c}{Response Acc}  &\multicolumn{2}{c}{AIME} &\multicolumn{2}{c}{MATH500}  &\multicolumn{2}{c}{LCB}& \multirow{2}{*}{Avg.}  \\ 
        &EN &TH &EN &TH &Lang &Think &EN &TH &EN &TH &EN &TH\\
        \midrule
        Sealion 70B Instruct &\textbf{89.5} &\textbf{78.2} &\textbf{9.056} &6.972 &90.0 &0.0& 20.0 &6.66 &69.8 &58.9&	35.4 &25.2& 52.8 \\
        Sealion 70B+SFT-v3+M3 &83.3 &78.0 &8.653 &\textbf{7.104}  &\textbf{90.4} &\textbf{100.0} &50.0 &\textbf{43.3} &\textbf{89.4} &\textbf{83.5} &59.4 &60.0&\textbf{74.6} \\
        Deepseek R1 70B &85.7 &74.3 &8.939 &6.329 &19.0 &84.2 &\textbf{63.3} &40.0 &88.4 &78.7& \textbf{64.7} &\textbf{62.8} &67.8 \\
    \bottomrule
    \end{tabular}
    \label{tab:sealion}
\end{table}

\section{Conclusion \& Limitation}
\label{conclusion}

In this work, we propose a method to enhance reasoning in language-specific models by combining two specialize models: one language-specific and another with long-thought reasoning capability. We showed that SFT \& merging can be a practical resources alternative for teaching model a reasoning capability, however due to combination of merging and SFT technique, it has certain limitations. Experimentally, we focus solely on merging DARE \citep{yu2024languagemodelssupermario} with a simple two-model setup and evaluate it on only one model family. Additionally, we do not optimize the instruction tuning subset, despite the availability of high-quality open-source instruction datasets such as Tulu3 \citep{lambert2025tulu3pushingfrontiers}. Furthermore, in this work, we focus on Thai as a case study; a broader multilingual investigation is left for future work.

At a higher level, several challenges remain in the realm of multilingual reasoning and model merging. These include the absence of culturally aware reasoning traces, performance disparities between low-resource and high-resource languages, and a limited understanding of the internal representations of reasoning in LLMs. Nonetheless, our goal is to advance LLMs in underrepresented languages, ensuring they remain competitive within the broader AI community.


\bibliography{iclr2025_conference}
\bibliographystyle{iclr2025_conference}

\appendix
\section{Appendix}

\subsection{Example of code-switching \& language accuracy problem}

\label{appendix:code_switching_example}

\begin{figure}[!ht]
\begin{center}
\includegraphics[width=\linewidth]{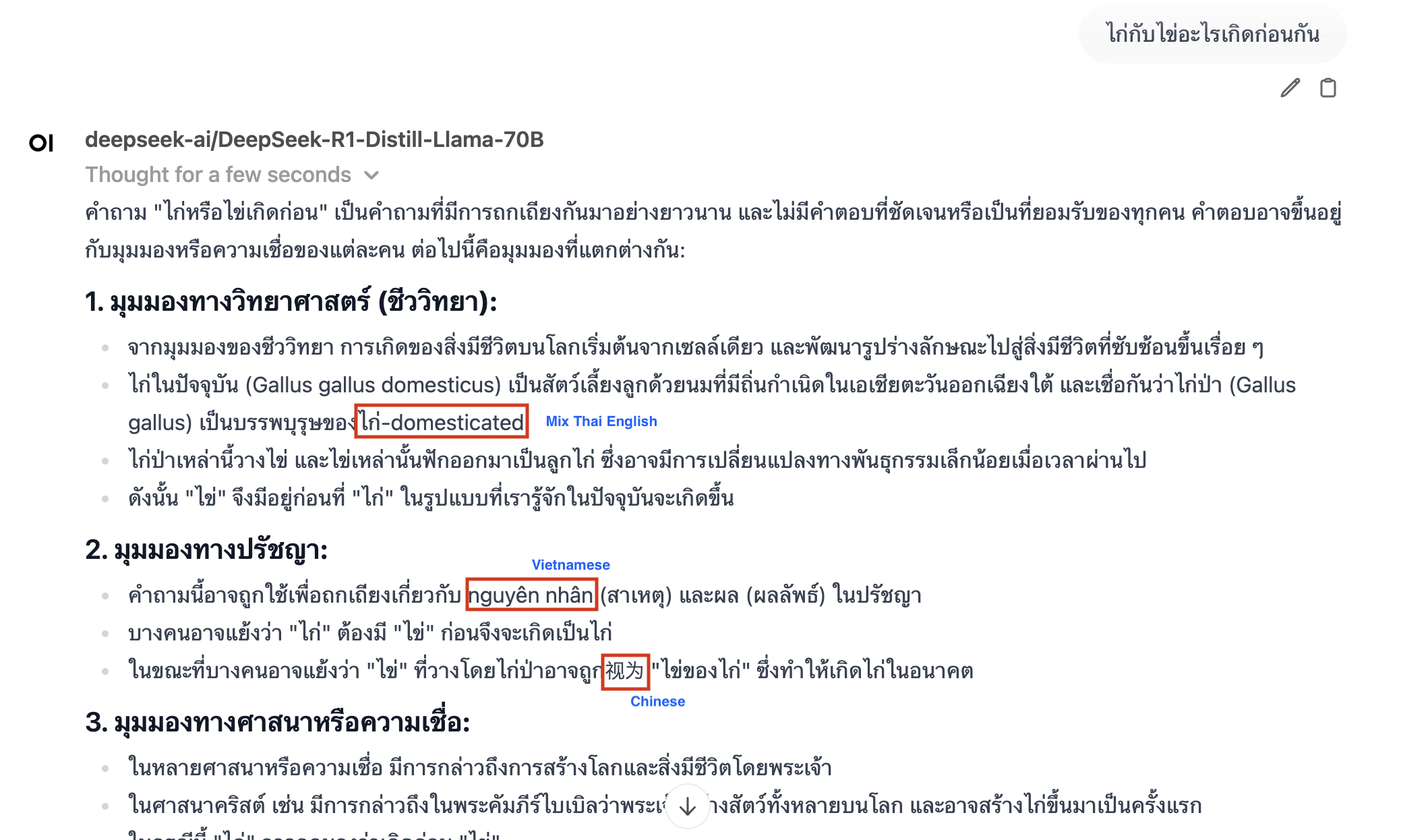}
\end{center}
\caption{Example demonstrate code-switching / language accuracy problem in DeepSeek R1 70B Distill. - The question is "Which came first, the chicken or the egg?" - The model generated a final response, but it was unsatisfactory as it contained unnatural code-switching that not in Thai.}
\label{fig:code_switch_problem}
\end{figure}

\begin{figure}[!ht]
\begin{center}
\includegraphics[width=\linewidth]{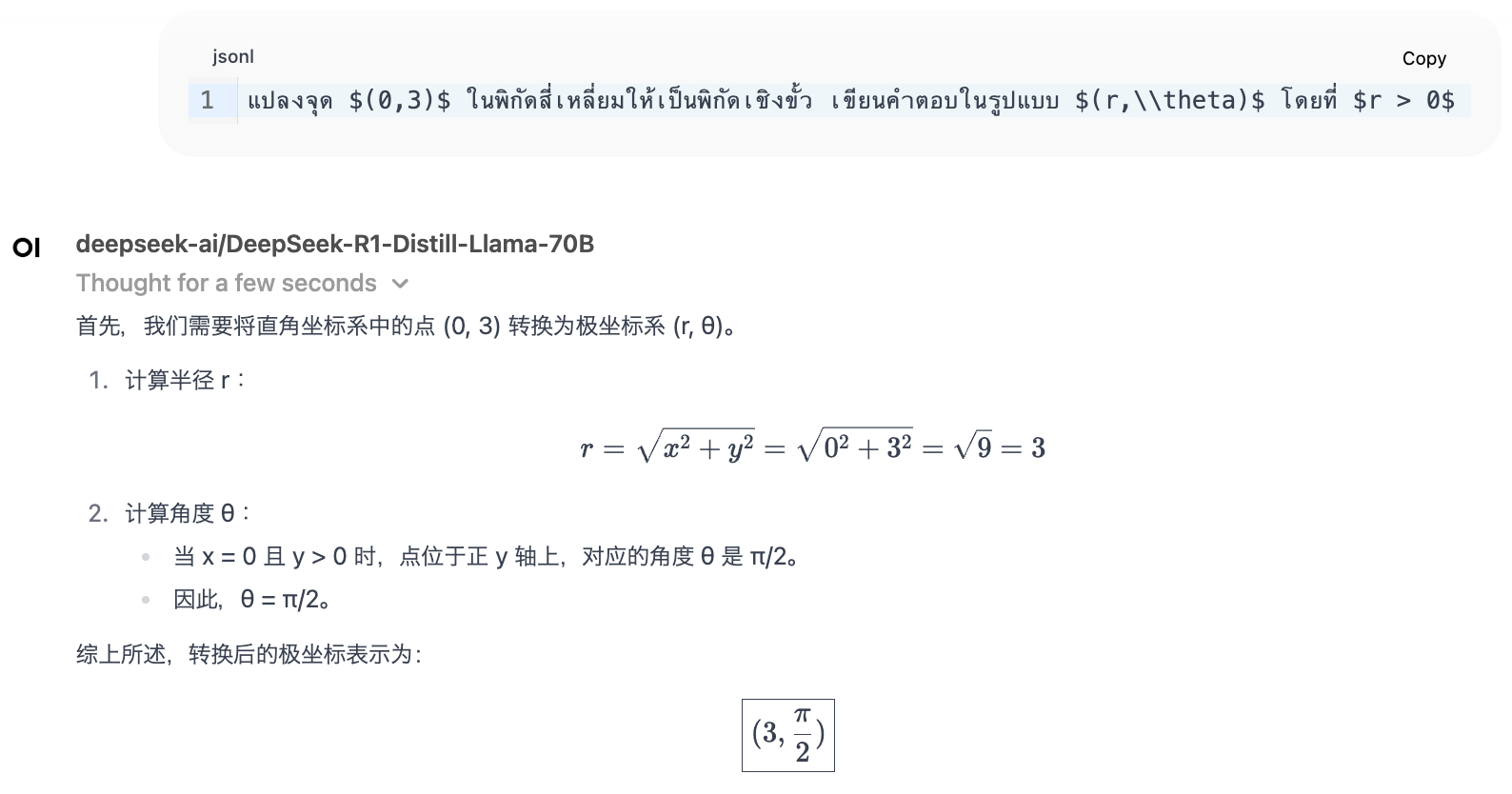}
\end{center}
\caption{Example demonstrate code-switching / language accuracy problem in DeepSeek R1 70B Distill. - The question is ``$\text{Convert the point } (0,3) \text{ in rectangular coordinates to polar coordinates.}$\\ \text{Enter your answer in the form } $(r,\theta), \quad \text{where} \quad r > 0, \quad 0 \leq \theta < 2\pi.$'' - The model generated a final response, but it was entirely in Chinese, which is not the usual language in Thai.}
\label{fig:code_switch2_problem}
\end{figure}

In Figure~\ref{fig:code_switch_problem} and Figure~\ref{fig:code_switch2_problem}, we demonstrate the problem more concretely. We show how code-switching manifests in real-world situations. First, there is code-switching, where the LLMs incorporate incorrect language words into the response. Second, the model ignores the given language order and responds in its familiar language.

\subsection{Language accuracy evaluation}
\label{appendix:language_acc}
In order to evaluate language accuracy, such as the example in Appendix~\ref{appendix:code_switching_example}
we focus on creating a verifiable rule that has two sub-rules:
\begin{enumerate}
    \item Valid responses should contain only Thai and English characters, excluding Chinese, Russian, or Vietnamese, as these represent languages commonly used by Thai speakers in daily life.
    \item English usage must follow native conventions, with the total number of English characters being fewer than Thai characters.
\end{enumerate}
The verifiable rule pseudo-code is shown in \ref{listing:lang_acc_impl}.

To ensure the validation works, we use prompts based on the 
\texttt{airesearch/WangchanThaiInstruct}~\citep{wangchaninstruct} \citep{wangchaninstruct} test set, due to its authenticity and the fact that it is the only Thai instruction dataset created by humans in Thai, making it representative of real prompts that a Thai person would write. 
To prevent unclear instructions, we explicitly instruct the LLM to generate responses based on the prompt language.We validate accuracy based only on rule adherence, not on the correctness of the answer itself—similar to IFEval \citep{ifeval}. Additionally, we performed non-greedy sampling at a temperature of 0.7 to simulate user scenarios where LLM responses is not static.

\begin{minipage}{\linewidth}
\begin{lstlisting}[frame=single, caption=Pseudo code for language accuracy validation function]
function is_mainly_thai(response: str):
    define thai_character_ranges
    define allowed_symbols (mathematical symbols, 
    Latin with diacritics, 
    quotes, punctuation, whitespace, emoji, etc.)

    initialize counters for thai_count, 
    english_count, 
    and other_chars set (chinese, russian, vietnamese, etc.)

    for each char in response:
        if char is Thai:
            increment thai_count
        elif char is in allowed_symbols:
            if char is English or Latin with diacritics:
                increment english_count
        else:
            add char to other_chars

    if other_chars is not empty:
        return False # contains non-Thai usage characters

    if english_count > thai_count:
        return False # content is not in Thai, but English

    return True

\end{lstlisting}
\label{listing:lang_acc_impl}
\end{minipage}

\subsection{Think Accuracy Evaluation}
\label{appendix:think_acc}

To evaluate think accuracy—the rate at which an LLM correctly utilizes the ability to think before generating a response—we define the problem as a verifiable rule for both the format and content of the thought process. Specifically, DeepSeek R1 uses the `\texttt{$<$think$>$}` and `\texttt{$</$think$>$}` tokens to separate its reasoning from the final solution. Our evaluation focuses on verifying whether the LLM-generated response meets the following criteria:

\begin{enumerate}
    \item \textbf{Does it follow the format?} – Does the response correctly include the `\texttt{$<$think$>$}` and `\texttt{$</$think$>$}` tokens?
    \item \textbf{Does it actually think?} – Our initial investigation revealed that DeepSeek R1 70B Distill, even when correctly formatting its response with `\texttt{$<$think$>$}` and `\texttt{$</$think$>$}`, sometimes generates an empty thought, such as 
    `\texttt{$<$think$>$}\textbackslash n\textbackslash n\texttt{$</$think$>$}`
\end{enumerate}

To apply this evaluation across various use cases, we enforce these verifiable rules on all responses generated across multiple benchmark datasets, including MT-Bench, IFEval, language-accuracy, AIME, MATH500, and LiveCodeBench. We then compute the accuracy based on the model's tendency to both format its thoughts correctly and generate non-empty reasoning.

The pseudocode for the verifiable rule implementation for think accuracy is provided in \ref{listing:think_acc_impl}.

\begin{minipage}{\linewidth}
\begin{lstlisting}[language=Python,frame=single, caption=Pseudo code for think accuracy validation function]
function is_think(response: str):
    if <think> or </think> in response:
        think_content = extract_content_between <think> and </think>
        if len(think_content.strip()) >= 0:
            return True
    return False

\end{lstlisting}
\label{listing:think_acc_impl}
\end{minipage}

\subsection{Merge config}
\label{appendix:merge_config}

To enhance understanding and transparency of our recipe, we provide our merge configuration for Mergekit below \citep{goddard2025arceesmergekittoolkitmerging}.

\begin{minipage}{\linewidth}
\begin{lstlisting}[frame=single, caption=Merge config: M1]
models:
  - model: meta-llama/Llama-3.1-70B
  - model: deepseek-ai/DeepSeek-R1-Distill-Llama-70B
    parameters:
      density: [0.3, 0.3, 0.3, 0.3]
      weight: [0.2, 0.2, 0.2, 0.2]
  - model: SFT-v1
    parameters:
      density: [1.0, 1.0, 1.0, 1.0]
      weight: [0.6, 0.6, 0.6, 0.6]
merge_method: dare_linear
base_model: meta-llama/Llama-3.1-70B
parameters:
  normalize: true
dtype: bfloat16

tokenizer:
  source: deepseek-ai/DeepSeek-R1-Distill-Llama-70B
\end{lstlisting}
\label{listing:M1}
\end{minipage}

\begin{minipage}{\linewidth}
\begin{lstlisting}[frame=single, caption=Merge config: M2]
models:
  - model: meta-llama/Llama-3.1-70B
  - model: deepseek-ai/DeepSeek-R1-Distill-Llama-70B
    parameters:
      density: [1.0, 1.0, 1.0, 1.0]
      weight: [0.6, 0.6, 0.6, 0.6]
  - model: SFT-v1
    parameters: 
      density: [0.3, 0.3, 0.3, 0.3]
      weight: [0.2, 0.2, 0.2, 0.2]
merge_method: dare_linear
base_model: meta-llama/Llama-3.1-70B
parameters:
  normalize: true
dtype: bfloat16
tokenizer:
  source: deepseek-ai/DeepSeek-R1-Distill-Llama-70B
\end{lstlisting}
\label{listing:M2}
\end{minipage}

\begin{minipage}{\linewidth}
\begin{lstlisting}[frame=single, caption=Merge config: M3]
models:
  - model: meta-llama/Llama-3.1-70B
  - model: deepseek-ai/DeepSeek-R1-Distill-Llama-70B
    parameters:
      density: [1.0, 1.0, 1.0, 0.3]
      weight: [0.6, 0.6, 0.6, 0.1]
  - model: SFT-v1
    parameters: 
      density: [0.3, 0.3, 0.3, 1.0]
      weight: [0.2, 0.2, 0.2, 0.7]
merge_method: dare_linear
base_model: meta-llama/Llama-3.1-70B
parameters:
  normalize: true
dtype: bfloat16

tokenizer:
  source: deepseek-ai/DeepSeek-R1-Distill-Llama-70B
\end{lstlisting}
\label{listing:M3}
\end{minipage}

\subsection{An example response from our model}
\label{appendix:example_from_model}

\begin{figure}[!ht]
\begin{center}
\includegraphics[width=\linewidth]{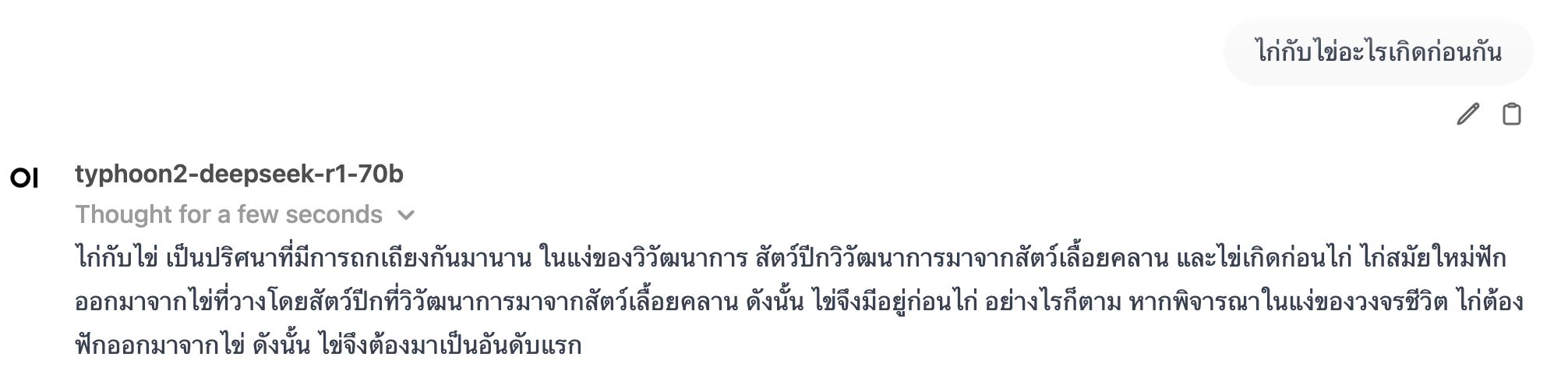}
\end{center}
\caption{Example from our model: The question is, 'Which came first, the chicken or the egg?' - The model successfully responds fully in Thai while reasoning through its thought process on general question.}
\label{fig:model_example_1}
\end{figure}

\begin{figure}[!ht]
\begin{center}
\includegraphics[width=\linewidth]{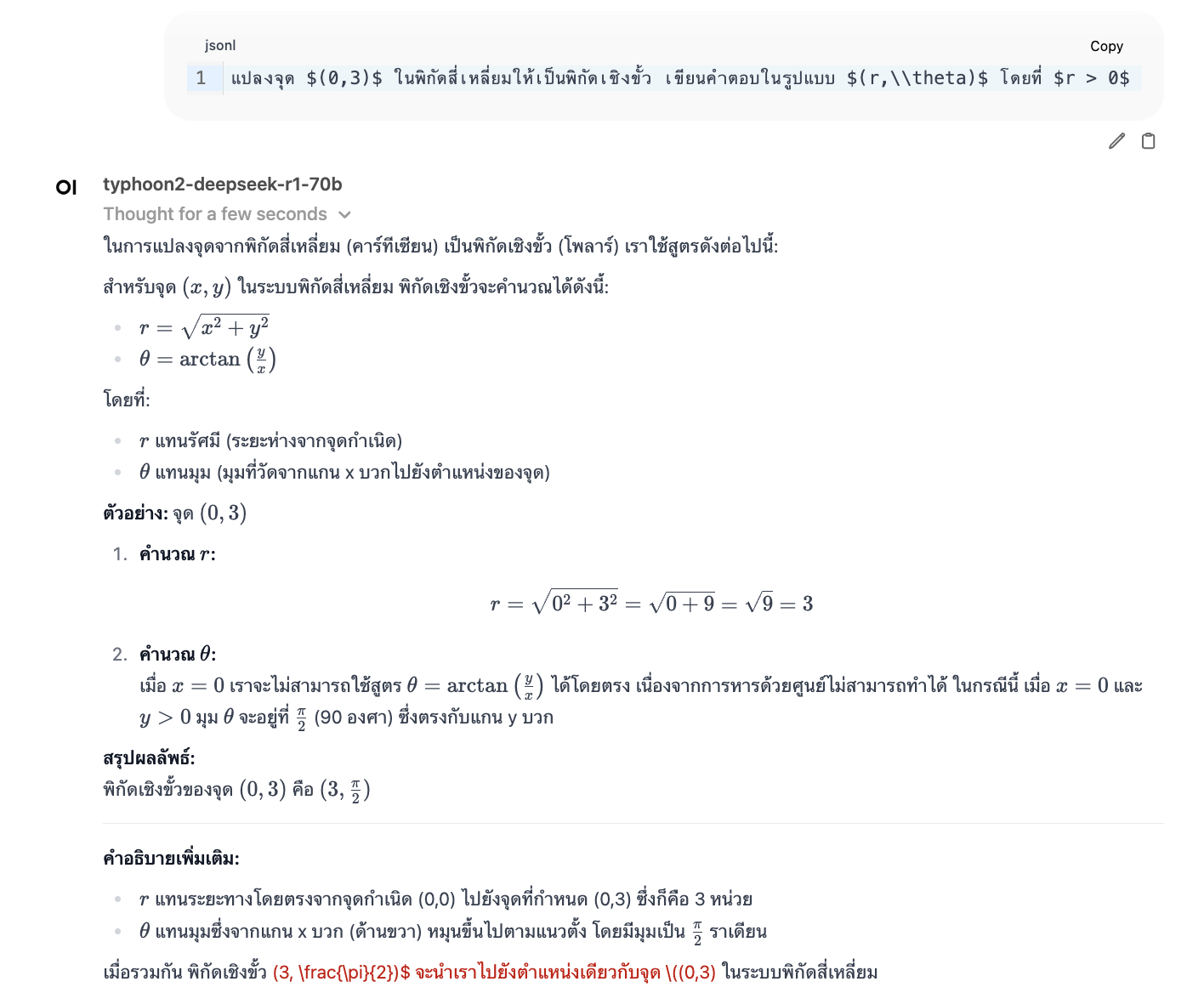}
\end{center}
\caption{Example demonstrate code-switching / language accuracy problem in DeepSeek R1 70B Distill. - The question is ``$\text{Convert the point } (0,3) \text{ in rectangular coordinates to polar coordinates.}$\\ \text{Enter your answer in the form } $(r,\theta), \quad \text{where} \quad r > 0, \quad 0 \leq \theta < 2\pi.$'' - The model successfully responds fully in Thai while reasoning through its thought process on math question.}
\label{fig:model_example_2}
\end{figure}

\end{document}

%% file: math_commands.tex
\usepackage{amsmath,amsfonts,bm}

\def\eqref#1{equation~\ref{#1}}

\def\1{\bm{1}}

\DeclareMathAlphabet{\mathsfit}{\encodingdefault}{\sfdefault}{m}{sl}
\SetMathAlphabet{\mathsfit}{bold}{\encodingdefault}{\sfdefault}{bx}{n}

